\pdfoutput=1
\documentclass[11pt]{article}

\usepackage[final]{acl}

\usepackage{times}
\usepackage{latexsym}
\usepackage{booktabs}
\usepackage{multirow}
\usepackage{caption}

\usepackage[T1]{fontenc}
\usepackage[utf8]{inputenc}
\usepackage{amsmath}

\usepackage{microtype}

\usepackage{inconsolata}

\usepackage{graphicx}

\title{Hidden in Plain Sight: Evaluation of the Deception Detection 

Capabilities of 
LLMs in Multimodal Settings}

\author{Md Messal Monem Miah\textsuperscript{1}, Adrita Anika\textsuperscript{2}\thanks{Work done outside of role at Amazon} , Xi Shi\textsuperscript{1}, Ruihong Huang\textsuperscript{1}\\ 
\textsuperscript{1}Texas A\&M University\\
\textsuperscript{2}Amazon\\
\normalsize{\texttt{\{messal.monem, xishi, huangrh\}@tamu.edu}},
\normalsize{\texttt{adritani@amazon.com}}} %\texttt{huangrh@cse.tamu.edu}}

\begin{document}
\maketitle

\begin{abstract}
Detecting deception in an increasingly digital world is both a critical and challenging task. In this study, we present a comprehensive evaluation of the automated deception detection capabilities of Large Language Models (LLMs) and Large Multimodal Models (LMMs) across diverse domains. We assess the performance of both open-source and proprietary LLMs on three distinct datasets—real-life trial interviews (RLTD), instructed deception in interpersonal scenarios (MU3D), and deceptive reviews (OpSpam). We systematically analyze the effectiveness of different experimental setups for deception detection, including zero-shot and few-shot approaches with random or similarity-based in-context example selection. Our findings indicate that fine-tuned LLMs achieve state-of-the-art performance on textual deception detection, whereas LMMs struggle to fully leverage multimodal cues, particularly in real-world settings. Additionally, we analyze the impact of auxiliary features, such as non-verbal gestures, video summaries, and evaluate the effectiveness of different prompting strategies, such as direct label generation and post-hoc reasoning generation. Experiments unfold that reasoning-based predictions do not consistently improve performance over direct classification, contrary to the expectations.
\end{abstract}
\section{Introduction}

Deception detection—the ability to identify intentionally misleading statements or behaviors—plays
a critical role in safeguarding security, justice, and
societal trust. Traditionally, its primary applications have been in criminalistics, particularly in interrogation and law enforcement settings such as suspect interrogations and security screenings. However, its relevance has expanded beyond these domains to border security~\citep{border}, healthcare~\citep{patient-doctor}, social media platforms~\citep{dd-social}, and consumer protection~\citep{opspam}. 

% In forensic and legal contexts, deception detection plays a pivotal role in assessing the credibility of witnesses and suspects, directly influencing investigative processes and judicial decisions~\cite{rltd, italian-dd}. Meanwhile, in e-commerce, fraudulent product reviews have become a growing concern, as deceptive reviews are used to manipulate consumer perception, inflate ratings, undermine trust in e-commerce platforms~\citep{li-etal-2013-identifying, opspam, fornaciari-poesio-2014-identifying}. 

% Despite its significance, human accuracy in detecting deception remains only slightly above chance—around 54\%~\citep{depaulo-54}—due to cognitive biases like truth bias~\citep{truth-bias} and the nuanced, context-dependent nature of deceptive cues. Advancing automated deception detection methods across these diverse domains is thus essential for mitigating fraud, preventing wrongful convictions, and preserving integrity in both digital and real-world interactions.

Despite its significance, deception detection remains inherently difficult, as human accuracy in detecting deception is only slightly above chance, $\sim54\%$~\citep{depaulo-54}. \textbf{Cognitive Load Theory}~\citep{vrij-clt} suggests that lying demands greater mental effort, which can lead to detectable inconsistencies, but deceivers often mitigate this by rehearsing or simplifying their fabrications~\citep{vrij2}. \textbf{Interpersonal Deception Theory}~\citep{buller-idt} highlights deception as an adaptive process, where deceivers adjust their behavior based on audience reactions, reducing the reliability of static detection methods. \citet{tdt} further explains humans’ bias toward assuming truthfulness, making them prone to overlooking deceptive cues. These challenges have driven the development of automated deception detection systems that systematically analyze linguistic, acoustic, and visual cues to improve reliability and scalability.

Researchers have increasingly explored automated approaches that combine advances in computer vision, natural language processing, and deep learning for deception detection. Early computational models in deception detection often relied on handcrafted features~\citep{Ekman2001, Zhang2020MultimodalDD, 8684170, Fan, Bai2019AutomaticLD}, drawing from facial micro-expressions, acoustic descriptors, and linguistic markers. With the emergence of deep learning, end-to-end architectures can directly learn deception-related patterns from raw multimodal data—text, audio, and video—leading to improved deception detection performance while reducing reliance on laborious feature engineering~\citep{dolos, SEPSIS, guo2024}. Despite these advances, existing deception detection systems still face challenges related to generalization, as deception cues vary across individuals, cultures, and contexts. Additionally, many deep learning models operate as black-box systems, making it unclear whether they genuinely capture deception-related patterns or rely on statistical shortcuts.

Recently, \textbf{Large Language Models (LLMs)} have demonstrated strong cognitive reasoning capabilities, excelling in tasks, such as emotion recognition~\citep{NEURIPS2024_c7f43ada, INSTERC, zhang-etal-2024-visual}, sentiment analysis~\citep{zhang-etal-2024-sentiment}, and fact verification~\citep{zhang-gao-2023-towards}. These models leverage large-scale pretraining and in-context learning to adapt to new tasks with minimal labeled data. LLMs’ ability to identify subtle linguistic cues, integrate multimodal inputs ~\citep{Chu2023QwenAudioAU, liu2023llava, zhang-etal-2023-video}, and and generate step-by-step reasoning behind the judgment through chain-of-thought prompting~\citep{cot} makes them promising candidates for automated deception detection. However, empirical evidence on LLM-driven deception detection, particularly in real-world multimodal settings, remains limited.

In this work, we take a comprehensive step toward filling this gap by challenging state-of-the-art LLMs with multiple deception detection tasks spanning three well-established datasets- \textbf{Real-life Trial Dataset, RLTD}~\citep{rltd}, \textbf{Miami University Deception Detection Database, MU3D}~\citep{mu3d}, and \textbf{Opinion Spam Dataset, OpSpam}~\citep{opspam}. These datasets cover deception across online, controlled, and real-world legal settings, collectively capturing diverse deception strategies and manifestations. The key contributions of this work are:

\begin{itemize}
    \item We benchmark several state-of-the-art open-source and proprietary LLMs for deception detection on three datasets, providing a large-scale comparison of these models on diverse deception detection scenarios. Additionally, We assess the performance of open-source large multimodal models on the two multimodal (RLTD, MU3D) datasets, offering insights into how visual and acoustic cues can impact deception detection performance.

    \item We explore various fine-tuning and inference setups, including zero-shot prompting, random and similarity-based example selection for few-shot learning. We further investigate how different prompting strategies (direct label generation vs. post-hoc reasoning generation) affect deception detection results, shedding light on the best strategies for designing LLM-driven deception detection pipelines.
    
    \item We incorporate additional features, such as non-verbal gestures for RLTD and video summaries for RLTD and MU3D, to evaluate the influence of auxiliary features on model performance.

\end{itemize}

By presenting a thorough empirical study of LLM-based deception detection across multiple domains and modalities, we contribute a holistic perspective on the efficacy and limitations of these models. 

% We present a large-scale benchmarking of state-of-the-art open-source and proprietary LLMs for deception detection across three datasets, covering diverse real-world scenarios. Our study further evaluates multimodal LLMs to assess the impact of visual and acoustic cues on deception detection. We explore various fine-tuning and inference strategies, including zero-shot prompting, random/similarity-based few shot example selection for in-context learning, and chain-of-thought reasoning, to identify optimal approaches for LLM-driven deception detection. Additionally, we incorporate non-verbal gestures and video summaries to analyze the role of auxiliary features in enhancing model performance. By conducting a comprehensive empirical study across multiple domains and modalities, we provide a holistic assessment of LLMs’ effectiveness and limitations in deception detection.
\section{Related Works}

Early research on automated deception detection leveraged handcrafted linguistic, syntactic, and lexical features, including Linguistic Inquiry and Word Count (LIWC) indicators, part-of-speech distributions, and n-gram features, to capture linguistic, psychological and stylistic patterns indicative of deception. These features were utilized in statistical models such as logistic regression, decision trees, and support vector machines (SVM) to classify deceptive and truthful statements~\citep{opspam, rltd, levitan-etal-2018-linguistic, Ekman2001, LIWC3, LIWC4}. Audio-based deception detection has relied on Mel-frequency cepstral coefficients (MFCCs) and prosodic cues, such as pitch and speaking rate, to distinguish deceptive from truthful speech~\citep{Hirschberg2005DeceptiveSpeech, levitan-etal-2018-linguistic, Bai2019AutomaticLD, BoL, Chebbi2021DeceptionDU}. Additionally, research in nonverbal deception detection has focused on facial Action Units (AUs) extracted from video data, which capture microexpressions and facial muscle movements associated with deceptive behavior~\citep{Ekman2001, fau1, Bai2019AutomaticLD, LIWC3, mathur2021affectawaredeepbeliefnetwork}. These approaches, though effective in constrained settings, often struggle with generalization across datasets and speaker variations, necessitating the exploration of more robust deep learning techniques.

Recent advances in deep learning have led to an increasing adoption of CNNs and LSTMs for deception detection tasks across both textual and multimodal domains~\citep{DL5, DL2, DL4, DL3, DL1}. Transformer based models and attention mechanisms have also been applied in recent deception detection research, leveraging contextual embeddings to capture subtle deception cues~\citep{tr1, tr2, tr3}. ~\citet{dolos} presents a novel method called Parameter-Efficient Crossmodal Learning (PECL) that uses a temporal adapter to capture temporal attention and a fusion module to merge audio and visual cues for audio-visual deception detection. Building on these developments, emerging research is now harnessing the capabilities of LLMs—whose success across diverse cognitive tasks underscores their potential—to capture intricate linguistic nuances and further enhance deception detection. In their study, \citet{LLM1} employ variants of the FLAN-T5 model~\citep{Flant5} to detect deception across a range of textual contexts. \citet{LLM2} investigates the effectiveness of LLMs in deception detection using a Retrieval Augmented Generation (RAG) framework for few-shot learning in various textual domains. Our work advances this line of research by investigating the application of LLMs in real-world multimodal scenarios.
\section{Background}

\subsection{Problem Definition}

Deception detection is the task of identifying whether a statement or behavior is deliberately misleading. We define this task as a binary classification problem, where the goal is to predict \( y \in \{\texttt{Truthful}, \texttt{Deceptive}\} \) given an input processed by a large language model. Formally, let \( p \) denote a task-specific prompt that instructs the model to process the input content and generate the classification label as either truthful or deceptive, and let \( t \) represent the textual content under analysis (for instance, a speech transcript or an online review). In the simplest setting, the input is
\(
x = p \,\odot\, t,
\)
where \(\odot\) denotes concatenation, and the model generates the prediction via
\(
y = f_\theta(x),
\)
where $f_\theta$ represents the LLM parameterized by $\theta$. 

Although textual cues can be highly informative for detection deception, additional cues may arise from non-verbal or multimodal sources. To account for such signals, we allow the input to be augmented by auxiliary features \( u \), which could include descriptive text of facial expressions and body movements, or a textual summary of the observed video content or speech characteristics. In that case, the model processes
\(
x = p \,\odot\, t \,\odot\, u.
\)
Furthermore, when employing large multimodal models (LMMs) with the capacity of handling audio or video, the input can incorporate raw audio or video directly, denoted by \( a \) and \( v \) respectively, such that
\(
x = p \,\odot\, t \,\odot\, [a, v]
\)

\subsection{Datasets}

\paragraph{\textbf{Real-life Trial Dataset (RLTD)}} ~\citet{rltd} is constructed from publicly available courtroom trial recordings. Labels are assigned based on trial outcomes, with guilty verdicts indicating deception and non-guilty verdicts or exoneration indicating truthfulness. In some cases, the same individual contributes both deceptive and truthful statements, capturing within-subject deception variations. The dataset includes 121 video clips (60 truthful and 61 deceptive) with transcripts. The videos are also annotated for non-verbal features using the MUMIN multimodal coding scheme~\citep{mumin}, focusing on facial expressions, gaze, head, and hand movements.

\paragraph{\textbf{Miami University Deception Detection Database (MU3D)}} ~\citet{mu3d} is a controlled deception dataset capturing instructed deception in interpersonal scenarios. Participants were asked to describe individuals they liked or disliked while alternating between truthfulness and deception. The dataset comprises 320 (160 truthful and 160 deceptive) videos with metadata, including trustworthiness ratings, anxiety ratings, demographic details, and full speech transcriptions.

\paragraph{\textbf{Opinion Spam Dataset (OpSpam)}} ~\citet{opspam} focuses on deception in online reviews and consists of 1600 reviews evenly split between truthful and deceptive opinions about hotels. Deceptive reviews were artificially generated by paid participants instructed to write persuasive but fabricated reviews, while truthful reviews were collected from genuine user feedback on platforms like TripAdvisor and Yelp. The dataset presents a linguistic deception challenge where fabricated narratives must be distinguished from authentic experiences.

Together, these datasets provide a rigorous benchmark for evaluating LLMs and LMMs in deception detection across legal, interpersonal, and online domains, ensuring a comprehensive assessment of their effectiveness.

\subsection{Baselines}
We evaluate the LLM based approaches against several deep-learning and transformer based baselines for text-only and multimodal deception detection. For the baselines, we extract modality-specific features using state-of-the-art pre-trained encoders. We obtain textual features from the final hidden states of the RoBERTa-base~\citep{roberta} model. For acoustic features, we use the final encoder hidden states of the Whisper-base~\citep{whisper} model, which has demonstrated robust performance in various audio tasks~\citep{miah, whisper2}. For visual features, we sample the input video at 30 fps and encode each frame using CLIP~\citep{CLIP}. 

We consider four baselines to compare against LLM-based approaches. First, we implement a text-only baseline by fine-tuning RoBERTa with a two-layer MLP classification head. Second, we follow ~\citet{bilstm} to employ a Bi-LSTM with attention network on the multimodal features described previously. We concatenate the resulting representations from each modality and use linear layers for multimodal classification. In unimodal scenarios, we simply predict the label from unimodal representations. Third, we follow  ~\citet{cnn1, DL4} to use CNN with global average pooling for feature encoding. Again, we concatenate the features across all modalities for multimodal deception detection. Finally, we replicate the Parameter-Efficient Crossmodal Learning (PECL) model proposed in  ~\citet{dolos}, which uses a 1D-convolution-based temporal adapter to learn modality-specific temporal attention alongside pre-trained Wav2Vec2 and ViT backbone models, supplemented by a Plug-in Audio-Visual Fusion (PAVF) module for crossmodal attention. This design enables PECL to achieve strong performance in the audio-visual setting. We conduct all experiments using stratified \textbf{4-fold cross-validation} across all three datasets.

\section{Experimental Setup}
\begin{table*}[ht]
\centering
\footnotesize
\begin{tabular}{l c c cc cc cc}  % <-- Added an extra 'c' for the new 'Config' column
\toprule
\multirow{2}{*}{\textbf{Model}} 
  & \multirow{2}{*}{\textbf{Config}} 
  & \multirow{2}{*}{\textbf{Modality}} 
  & \multicolumn{2}{c}{\textbf{RLTD}} 
  & \multicolumn{2}{c}{\textbf{MU3D}} 
  & \multicolumn{2}{c}{\textbf{OpSpam}} \\
  \cmidrule(lr){4-5}\cmidrule(lr){6-7}\cmidrule(lr){8-9}
& & & \textbf{Acc} & \textbf{F1} & \textbf{Acc} & \textbf{F1} & \textbf{Acc} & \textbf{F1} \\
\midrule
\multicolumn{9}{c}{\textbf{Baselines}} \\  % <-- New row with "Baselines" centered
\midrule
RoBERTa-ft 
  & - & t        & 76.31 & 76.22 & \textbf{67.92} & \textbf{67.81} & 88.10 & 88.09 \\
\midrule
\multirow{4}{*}{BiLSTM+Attention}
  & \multirow{4}{*}{-} & t        & 69.42 & 69.37 & 65.94 & 65.76 & \underline{90.45} & \underline{90.45} \\
  &                    & a        & 68.04 & 68.02 & 62.19 & 61.82 & -     & -     \\
  &                    & v        & 75.48 & 75.38 & 55.21 & 55.11 & -     & -     \\
  &                    & t, a, v  & 77.14 & 77.04 & 62.29 & 62.19 & -     & -     \\
\midrule
\multirow{4}{*}{CNN}
  & \multirow{4}{*}{-} & t        & 64.46 & 64.39 & 64.06 & 63.91 & 86.37 & 86.36 \\
  &                    & a        & 60.88 & 60.13 & 60.42 & 60.26 & -     & -     \\
  &                    & v        & \underline{82.09} & \underline{82.09} & 54.06 & 53.95 & -     & -     \\
  &                    & t, a, v  & \textbf{83.47} & \textbf{83.44} & 60.41 & 60.38 & -     & -     \\
\midrule
PECL 
  & - & a, v      & 80.17 & 80.13 & 56.56 & 56.56 & -     & -     \\
\midrule
\multicolumn{9}{c}{\textbf{LLM Inference}} \\
\midrule
%--- Single-row LLMs (Few shot, t) ---
LLaMA 3.1 & Few shot & t & 71.69 & 77.11 & 57.03 & 56.84 & 62.93 & 62.47 \\
\midrule
Gemma 2   & Few shot & t & 71.69 & 71.38 & 55.08 & 53.75 & 64.81 & 64.51 \\
\midrule
GPT-4o    & Few shot & t & 79.55 & 79.49 & 55.70 & 53.87 & 74.50 & 74.00 \\
\midrule
%--- Multi-row LLMs (Zero shot) ---
\multirow{2}{*}{LLaVA-NEXT-Video} & \multirow{2}{*}{Zero shot} & v     & 52.06 & 43.55 & 50.00 & 33.33 & - & - \\
&  & t, v  & 64.46 & 61.79 & 50.31 & 50.31 & - & - \\
\midrule
\multirow{2}{*}{Qwen2VL}          & \multirow{2}{*}{Zero shot} & v & 51.24 & 37.95 & 50.31 & 39.94 & - & - \\
&  & t, v  & 63.64 & 60.32 & 52.50 & 52.35 & - & - \\
\midrule
\multirow{2}{*}{MERaLiON-AudioLLM}& \multirow{2}{*}{Zero shot} & a     & 66.94 & 66.11 & 49.06 & 33.97 & - & - \\
&  & t, a  & 66.12 & 63.57 & 49.38 & 34.12 & - & - \\
\midrule
\multicolumn{9}{c}{\textbf{LLM Finetuning}} \\
\midrule
\multirow{2}{*}{LLaMA 3.1} & \multirow{2}{*}{-} & t & 69.63 & 69.62 & 57.74 & 57.58 & \textbf{92.25} & \textbf{92.24} \\
&                          & t, u   & 72.72 & 72.46 & - & - & - & - \\
\midrule
\multirow{2}{*}{Gemma 2}   & \multirow{2}{*}{-} & t     & 75.21 & 75.19 & \underline{66.56} & \underline{66.55} & 90.25 & 90.18 \\
&                          & t, u   & 75.21 & 75.17 & - & - & - & - \\
\midrule
\multirow{2}{*}{Qwen2VL}   & \multirow{2}{*}{-} & v     & 57.85 & 57.10 & 52.20 & 51.43 & - & - \\
&                          & t, v   & 71.90 & 71.90 & 56.25 & 53.51 & - & - \\
\bottomrule
\end{tabular}
%\captionsetup{font=scriptsize}
\captionsetup{font=footnotesize}

%\caption{Comparison of Baselines and LLM-Based Evaluations Across Modalities (t: text, a: audio, v: video, u: non-verbal features)}
\caption{Comparison of Baselines and LLM Results Across Modalities (t: text, a: audio, v: video, u: non-verbal features)}

\label{tab:myresults}
\end{table*}

% In this section, we outline the experimental framework used to evaluate the deception detection capabilities of LLMs and LMMs. We compare these generative models against traditional deep learning and transformer based baselines, employ multiple prompting strategies, and explore different few-shot example selection techniques. We conduct all experiments using stratified \textbf{4-fold cross-validation} across all three datasets.

We evaluate three Large Language Models (LLMs) for their deception detection capabilities: LLaMA3.1-8B~\citep{llama3.1}, Gemma2-9B~\citep{gemma2}, and GPT-4o~\citep{gpt4o}. Additionally, we assess the performance of various Large Multimodal Models (LMMs), categorized based on their modality specialization. For video-language models, we consider LLaVA-NEXT-Video~\citep{llavanextvideo} and Qwen2VL~\citep{qwen2vl}, while MERaLiON-AudioLLM~\citep{meralion} and Qwen2-Audio~\citep{qwen2audio} serve as the audio-language models. These models represent state-of-the-art architectures in language and multimodal understanding, offering a diverse perspective on deception detection across textual, audio, and visual modalities.

\subsection{Experimental Configurations}

We evaluate both zero-shot and few-shot inference setups. In zero-shot evaluation, the model receives only a task description prompt and input data without labeled examples. In few-shot evaluation, the model is provided with a set of labeled examples for in-context learning. Specifically we have experimented with \( n = \{2, 4, 6, 8, 10\} \), as number of in-context examples. Under the zero-shot and few-shot setups, we experiment with various strategies and configurations, outlined below.

\subsubsection{Response Generation Strategies}
To systematically assess deception detection performance, We investigate two different response generation strategies: \textbf{direct label prediction}, where the model directly generates the label for the input as either \texttt{Truthful} or \texttt{Deceptive} without additional reasoning, and \textbf{post-hoc reasoning generation}, where the model is prompted to first generate the classification label \( y \) and then provide a justification \( r \), such that: \(
(y, r) = f_{\theta}(x) \),  where \( x \) is the input and \( f_{\theta} \) represents the model parameterized by \( \theta \). The generated reasoning \( r \) serves as a justification for the classification decision, allowing for better interpretability of deception detection outcomes. We also evaluate the chain-of-thought prompting for reasoning generation. However, post-hoc reasoning generation is eventually adapted for better performance and interpretability, with further analysis provided in Appendix~\ref{cot}.

\subsubsection{In-Context Example Selection Strategies}
For the few-shot prompting setup, we explore different strategies for selecting in-context examples. Similar to the baselines, we employ a 4-fold split for in-context example selection. The random selection approach involves choosing an equal number of truthful and deceptive examples randomly from the other 3 splits. In contrast, the similarity-based selection methods employ sentence-transformers to encode the target input and dataset samples, allowing for similarity-based retrieval. Within this method, we examine two variants: \textbf{sim-top}, which selects the most similar examples irrespective of their label, and \textbf{sim-pair}, which ensures a balanced selection of truthful and deceptive examples based on similarity ranking.

\subsubsection{Auxiliary Features}
We incorporate additional auxiliary features on top of the textual contents in the multimodal datasets, that provide valuable non-verbal and contextual information. As a first set of features for the RLTD dataset, we include a curated selection of 16 non-verbal features, capturing facial expressions and body movements indicative of deceptive behavior. The features names are listed in Appendix \ref{non-verbal}.  These features allow the model to leverage fine-grained behavioral cues that are often imperceptible in textual analysis alone. In addition to non-verbal gestures, we experiment with video and audio summaries as auxiliary inputs. A video-language model, LLaVA-NeXT-Video is employed to generate summaries of the visual content, extracting key information regarding speaker posture, facial expressions, and body movements indicative of stress or deception. Similarly, an audio-language model, Qwen2-Audio is used to summarize the tonal and acoustic features of the speech, identifying variations in pitch, intonation, and vocal stress patterns. These summaries provide a higher-level contextual representation of the non-verbal elements within the dataset, aiding in deception detection by supplying a multimodal understanding of deceptive cues for the RLTD and MU3D datasets.

\subsubsection{Fine-Tuning}
To further enhance model performance, we fine-tune open-source LLMs using the LLaMA-Factory~\citep{llamafactory} framework. We specifically fine-tune LLaMA3.1-8B, Gemma2-9B, and Qwen2-VL-7B. This fine-tuning process allows the models to better adapt to the nuances of deception detection by learning from domain-specific patterns and optimizing their ability to process multimodal cues effectively.

\section{Results \& Analysis}
\begin{table*}[ht]
    \centering
    \small
    % \renewcommand{\arraystretch}{1.2}
    % \resizebox{\columnwidth}{!}{
    \begin{tabular}{llcccc}
        \toprule
        \multicolumn{2}{c}{} & \multicolumn{4}{c}{\textbf{Cues}} \\
        \cmidrule(lr){3-6}
        \textbf{LLM} & \textbf{Dataset} & \textbf{Details} & \textbf{Vagueness} & \textbf{Filler Words} & \textbf{Justification} \\
        \midrule
        LLaMA 3.1 & RLTD & 86.20\% (29) & 78.87\% (14) & 84.00\% (25) & 43.37\% (23) \\
        & MU3D & 57.70\% (52) & 31.25\% (16) & 60.7\% (28) & 69.23\% (13) \\
        & OpSpam & 65.26\% (1091) & 63.25\% (117) & - & 42.10\% (27) \\
        \midrule
        Gemma 2 & RLTD & 76\% (25) & 63.63\% (22) & 83.33\% (12) & 100\% (7) \\
        & MU3D & 75.0\% (8) & 66.67\% (9) & 50\% (4) & 62.5\% (8) \\
        & OpSpam & 68.88\% (50) & 70.83\% (24) & - & - \\
        \midrule
        GPT-4o & RLTD & 73.07\% (26) & 80.0\% (20) & 66.67\% (3) & 85.71\% (7) \\
        & MU3D & 61.90\% (21) & 100\% (2) & 100\% (1) & 100\% (3) \\
        & OpSpam & 58.75\% (80) & 77.77\% (9) & - & - \\
        \bottomrule
    \end{tabular}
    % }
    \captionsetup{font=footnotesize}
    \captionsetup{justification=centering}
    \caption{Accuracy percentages for different models and cues. The number of total data points is in paranthesis.}
    \label{tab:cues}
\end{table*}
In Table~\ref{tab:myresults}, we focus on the best configurations for LLM inference across RLTD, MU3D, and OpSpam, leaving a more detailed analysis to subsequent sections. While, the text-only LLMs, GPT-4o, LLaMA 3.1, and Gemma 2, manage to narrow some of the gap with the baselines on RLTD and MU3D datasets, their few-shot configurations do not consistently outperform the strongest baselines. GPT-4o reaches an F1 score of 79.49 on RLTD and 74.00 on OpSpam, signaling modest gains over other LLMs in the few-shot setup. On the contrary, zero-shot variants of LLaVA-NEXT-Video and Qwen2VL on RLTD and MU3D datasets remain less effective, especially when relying solely on video features, indicating a limited capacity to exploit visual cues without additional training. Even in the multimodal setup, they fail to surpass the CLIP-based video-only baselines. A similar pattern emerges for MERaLION-AudioLLM, which exhibits moderate zero-shot performance on RLTD using audio or multimodal inputs, yet still lags behind the Whisper-based audio-only baselines. These results suggest that LMMs fail to extract necessary cross-modal information for deception detection, unlike their multimodal baseline counterparts.

When fine-tuned, LLaMA 3.1 achieves state-of-the-art performance on the OpSpam dataset. Additionally, fine-tuning using non-verbal features boosts performance over just using the transcripts for RLTD. Gemma 2 raises its MU3D F1 score to 66.55 and achieves 90.18 F1 on OpSpam. Likewise, Qwen2VL experiences a performance boost on RLTD once text and video features are fine-tuned jointly. Nevertheless, even these tailored LLMs do not consistently match or surpass the strongest baselines for the multimodal datasets. 

\subsection{Comparison of CNN Baselines and vision LLMs}
Experimental results demonstrate that the CNN baselines perform the best when video features are used alone or fused with text and audio, underscoring the importance of visual information on RLTD’s unrehearsed deception, where deception-related micro-expressions and body movements are depicted in the video. By contrast, MU3D contains scripted deception, enabling actors to mask deception-related acoustic and visual cues while they are on record. As a result, fine-tuned RoBERTa and Gemma-2 outperform CNNs on this dataset. This observation also explains why text-only LLMs underperform compared to multimodal CNN baselines on RLTD, as they cannot utilize the nuanced visual and acoustic cues. 

During training, CNNs learn to align and fuse temporal cues across modalities, allowing them to attend to deception-relevant patterns like micro-expressions and movement trajectories. By contrast, vision-language models like LLaVA-NEXT-Video and Qwen2VL rely on zero-shot pre-training, focused on captioning, object tracking, and OCR, and thus lack inherent deception-specific cognitive knowledge. Inspection of their generated video summaries further reveals why they miss critical deception cues. An example video summary from LLaVA-NEXT-Video using the prompt presented in Appendix~\ref{subsec:vid-sum-prompt} - \textit{In the video, a woman is seated at a table, wearing glasses and a red blouse, engaging in a conversation or an interview. Her facial expressions are calm and composed, with minimal micro-expressions, and her eye movements are steady, suggesting a controlled demeanor. Her body language is relaxed, with minimal hand gestures and head movements, indicating a composed and collected demeanor. There are no visible stress signs or fidgeting patterns, and her posture remains consistent throughout the video...}

It is evident from the generated summary that the vision–language models such as LLaVA-NEXT-Video describe the scenes and the objects well, yet they consistently miss the fine-grained behavioural cues annotated in the dataset, e.g. \textit{raised eyebrows, gaze at interlocutor, downward lip movement, repeated nods, bilateral hand movement, complex hand trajectories} for this particular video. Consequently, these LMMs often report contradictory observations (e.g., `minimal hand gestures') where, in fact complex hand movements are present in the video. They trail CNN baselines even after fine-tuning on transcripts and video. A key reason is their limited temporal resolution: Qwen2VL is pretrained at \textbf{2 fps}, and LLaVA-NEXT recommends \textbf{16 frames per video}, whereas our CNN baselines operate on \textbf{10 fps} streams, capturing and tracking subtler micro-expressions. Raising the frame rate for LMMs increases latency and GPU memory requirements, curbing scalability. Taken together, the performance gap of CNN baselines and large vision-language models reflects domain-specific temporal limitations, pre-training biases, and practical resource constraints of current LMMs.

\subsection{Interpreting LLMs' Reasoning}
% \documentclass{article}
% \section{Analysis of LLM's Reasoning}

%\begin{table}[h]

Table~\ref{tab:response_generation_shots} in Appendix~\ref{reasoning gen} shows that direct label prediction and post-hoc reasoning generation often lead to similar performance. In RLTD, generating reasoning lowers performance across all models. However, for MU3D and OpSpam, we occasionally observe some improvements when reasoning is generated. Considering marginal and occasional gains from generating reasoning and associated additional costs, we adopt direct label prediction for further experiments. However, reasoning remains valuable for understanding the LLM’s decision-making, helping to identify biases and patterns in deception detection. We analyze both correctly classified and misclassified instances, examining patterns based on linguistic cues to understand LLMs strengths and limitations.

\paragraph{Specificity and Detail.}
To quantify the use of  \textit{specificity and detail} as a cue for deception detection, we identified instances where the model explicitly referenced `specific detail' in its reasoning and assessed accuracy based on correctly classified samples. As shown in Table~\ref{tab:cues}, models consistently used this cue, with accuracy ranging from 57\% to 86\%.  Notably, for the RLTD dataset, which consists of courtroom trials, accuracy was higher across all three models. This suggests that specific details are more informative in legal contexts, where testimonies often contain detailed accounts of events, locations, and actions, making specificity a stronger indicator of truthfulness. 
% In contrast, MU3D, which involves interpersonal deception related to emotions, and OpSpam, which consists of deceptive hotel reviews, present different challenges. 
Emotional deception, as in MU3D, may not always involve factual inconsistencies, making reliance on details less effective. Similarly, in the case of online reviews, deceptive reviewers can fabricate highly detailed experiences, while genuine reviewers may provide concise feedback without elaborate narratives.
To further investigate this behavior, we analyzed 86 randomly selected OpSpam samples where the LLaMA model referenced specificity in its reasoning. Of these, 67 lacked detail and were all classified as deceptive, misclassifying 13 truthful reviews. In contrast, 19 were classified as truthful due to specific details, yet 7 were actually deceptive (Figure \ref{fig:llm_reason2} \textit{Example 12}). This bias toward treating specificity as a truth cue aligns with Reality Monitoring Theory (RMT), which links truthfulness to sensory-rich statements ~\citep{vrij2008detecting}. However, in online reviews, deceptive writers may create vivid narratives, while truthful reviewers might be concise. This over-reliance on specificity exposes a key limitation of LLM’s reasoning process.

\paragraph{Vagueness.}
We examine the models' reliance on \textit{vagueness} as a deception cue. Table ~\ref{tab:cues} shows LLMs consistently use this cue, with GPT-4o demonstrating the highest accuracy. Analyzing LLaMA’s behavior, we found that in MU3D, all 16 vagueness-based classifications were deceptive, misclassifying 11 truthful cases,  suggesting that the model struggles to distinguish between genuine uncertainty and deceptive ambiguity in interpersonal communication. In RLTD, 14 instances were flagged as deceptive, with 11 correctly classified, indicating a slightly better alignment with deception patterns in courtroom testimonies. In OpSpam, 92 of 117 flagged cases were classified as deceptive (63.25\% accuracy). This bias toward associating vagueness with deception often leads to overgeneralization and misclassification as vagueness can naturally occur in truthful statements due to memory recall limitations or subjective expression. For instance, in MU3D example (Appendix, Figure~\ref{fig:llm_reason2} \textit{Example 7}), the speaker expresses strong negative emotions about a peer, saying, “He’s gotten my friends in trouble,” and “we stopped hanging out with him just because the, that whole reason,” without clearly specifying what “that whole reason” entails. This conversational vagueness i.e. the use of non-specific phrases led the model to classify the statement as deceptive. This misclassification highlights how the model may over-rely on surface-level ambiguity as a deception signal, failing to account for the emotional and informal nature of interpersonal speech. In emotionally charged dialogue, vague references can reflect genuine uncertainty or conversational style rather than intent to deceive.

\paragraph{Hesitation and Filler Words.}
We investigate LLM’s reliance on \textit{hesitation and filler words} as deception cues. LLMs frequently associates verbal disfluencies (e.g., `uh,' `um') with deception, aligning with Cognitive Load Theory ~\citep{vrij2008detecting}, which suggests that lying requires greater mental effort, leading to pauses and hesitations. As shown in Table ~\ref{tab:cues}, GPT-4o relies on filler words less compared to LLaMa 3.1 and Gemma 2. Note, this cue was not used in OpSpam, as it comprises written reviews. We find that reliance on this cue sometimes leads to correct classifications—such as in Figure \ref{fig:llm_reason1} \textit{Example 1, 2}, where hesitation appeared alongside vagueness or contradictions. However, misinterpretations also occur, as seen in Figure \ref{fig:llm_reason1} \textit{Example 3}, where hesitation in a truthful statement resulted in a false deception label. Hesitation paired with detailed responses is often assumed to indicate truthfulness, correctly classified in Figure \ref{fig:llm_reason1} \textit{Example 4} but misapplied in \textit{Example 5}. 
% These findings suggest that while hesitation and specificity can be useful cues, the model needs a more nuanced understanding of context to reduce systematic errors.

\paragraph{Justification.}
 To assess the LLM’s use of \textit{justification} as a cue, we identified instances where `justify,' `justifies,' or `justification' appeared in its reasoning and reported accuracy in Table ~\ref{tab:cues}. The LLM often links justifications and indirect answers to deception, aligning with Criteria-Based Content Analysis (CBCA) ~\citep{vrij2008detecting}, which associates evasiveness with deception. Gemma applied this cue effectively in RLTD, correctly classifying 6 out of 7 cases. However, in MU3D, it consistently associated justification with deception, predicting all 8 instances as deceptive with 62.5\% accuracy. This suggests the model struggles to differentiate between genuine explanations and intentional deflection.

\paragraph{Emotions.}
We analyzed how LLMs interpret \textit{emotions} in deception detection, finding that they often associate strong emotional reactions with truthfulness. This aligns with Statement Validity Analysis (SVA) ~\citep{vrij2008detecting}, which considers spontaneous emotions a sign of genuine experiences. While this assumption sometimes led to correct classifications (Figure \ref{fig:llm_reason2} \textit{Example 10}), it also resulted in misclassifications. For instance, the model mistakenly labeled a deceptive statement as truthful when exaggerated emotions were used to appear credible (Figure \ref{fig:llm_reason1} \textit{Example 6}) and failed to recognize playful language, misinterpreting emotional expression (Figure \ref{fig:llm_reason2} \textit{Example 11}). In MU3D, genuine expressions of admiration and affection were frequently misclassified as deception (Figure \ref{fig:llm_reason2} \textit{Example 8, 9}). This indicate that LLM often lacks the ability to accurately interpret emotions.

\subsection{Impacts of In-context Examples}
%\begin{table}[ht]
\begin{table}[t]
\centering
\small  % or \footnotesize
\resizebox{\columnwidth}{!}{
\begin{tabular}{l c cc cc cc}
\toprule
% -- First header row --
\textbf{LLM} 
  & \textbf{Example}
  & \multicolumn{2}{c}{\textbf{RLTD}}
  & \multicolumn{2}{c}{\textbf{MU3D}}
  & \multicolumn{2}{c}{\textbf{OpSpam}} \\
\cmidrule(lr){3-4}\cmidrule(lr){5-6}\cmidrule(lr){7-8}
% -- Second header row --
 & \textbf{selection}
 & \textbf{Acc} & \textbf{F1}
 & \textbf{Acc} & \textbf{F1}
 & \textbf{Acc} & \textbf{F1} \\
\midrule

% ---- Body of table ----
\multirow{3}{*}{LLaMA 3.1} 
  & \textit{random}     & 68.87 & 68.14 & 51.72 & 51.18 & 59.62 & 59.19 \\
  & \textit{sim-pair} & \underline{71.69} & \underline{71.11} & 54.76 & 54.14 & 58.04 & 57.78 \\
  & \textit{sim-top}   & 71.28 & 70.25 & \textbf{57.03} & \textbf{56.84} & \underline{62.93} & \underline{62.47} \\
\midrule

\multirow{3}{*}{Gemma 2} 
  & \textit{random}    & 69.63 & 69.52 & 54.22 & 52.34 & 57.70 & 57.59 \\
  & \textit{sim-pair} & \underline{71.69} & \underline{71.38} & 54.92 & 53.69 & 60.14 & 59.98 \\
  & \textit{sim-top}   & 71.07 & 70.68 & \underline{55.08} & \underline{53.75} & \underline{64.81} & \underline{64.51} \\
\midrule

\multirow{3}{*}{GPT-4o} 
  & \textit{random}    & 71.69 & 71.39 & 53.20 & 46.86 & 68.40 & 67.58 \\
  & \textit{sim-pair} & \textbf{79.55} & \textbf{79.49} & 55.08 & 50.35 & 71.87 & 71.23 \\
  & \textit{sim-top}   & 77.69 & 77.69 & \underline{55.70} & \underline{53.87} & \textbf{74.50} & \textbf{74.00} \\
\bottomrule
\end{tabular}
}
\captionsetup{font=scriptsize}
\caption{Performance comparison of example-selection strategies across RLTD, MU3D, and OpSpam. The best overall results are in \textbf{bold}, while model-specific best performances are \underline{underlined}.}
\label{tab:example_selection}
\end{table}

Table~\ref{tab:example_selection} presents a comparison of three in-context example selection strategies—random, sim-top, and sim-pair under a few-shot prompting setup. In general, both similarity-based methods (sim-pair and sim-top) surpass random selection, demonstrating the importance of carefully curating in-context examples. The principal distinction between sim-top and sim-pair lies in label balancing: sim-top selects the most similar examples regardless of their labels, whereas sim-pair enforces a balanced set of truthful and deceptive instances among those most similar. In terms of the LLMs, GPT-4o exhibits the highest average improvement (7.18\% F1 score) when similarity-based few-shot examples are introduced, demonstrating more robust in-context learning capabilities relative to LLaMA3.1-8B and Gemma2-9B. 

\begin{figure}
    \centering
    \includegraphics[width=0.35\textwidth]{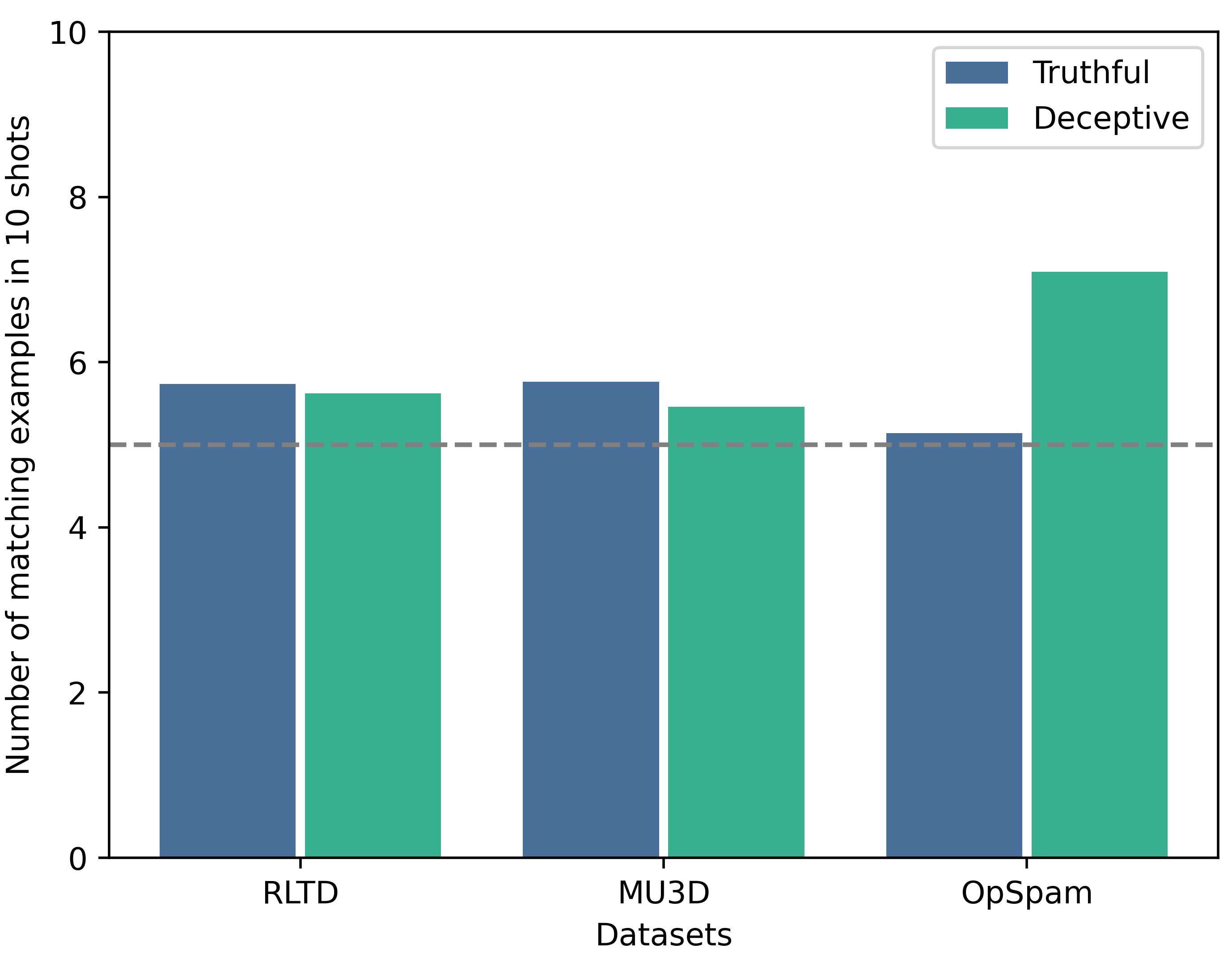}
    \captionsetup{font=scriptsize}
    \caption{Average number of matching examples in 10-shot for sim-top strategy.}
    \label{fig:num_example}
\end{figure}
Looking closely at RLTD, the sim-pair approach slightly outperforms sim-top. In contrast, on MU3D and OpSpam, sim-top provides superior results, particularly on OpSpam, where the average F1 score improvement, $\sim4\%$, is notably higher than that observed on MU3D $\sim2\%$. Figure~\ref{fig:num_example} further illuminates these findings by illustrating the average number of `matching' examples (i.e., truthful examples for a truthful query and deceptive examples for a deceptive query) retrieved in a 10-shot setup using sim-top. For RLTD and MU3D, this number hovers around five, effectively mirroring the label balance of sim-pair. However, in OpSpam, especially for the deceptive queries, the average number of retrieved deceptive examples rises to about 7.1, enabling a substantial boost in performance. Concretely, this increase in label-specific examples elevates the deceptive-class F1 score from approximately 67\% under sim-pair to 73\% under sim-top. This finding also suggests that deceptive reviews in the OpSpam dataset exhibit a higher degree of semantic similarity compared to the other datasets, hence easily identifiable by the retriever. This OpSpam dataset specific bias and the implications are further discussed in Appendix~\ref{sec:opspam}.

\subsection{Impacts of Auxiliary Features}

\begin{table}[ht]
\centering
\small
\resizebox{\columnwidth}{!}{
\begin{tabular}{l c c cc cc}
\toprule
\textbf{LLM} & \textbf{Feats} & \textbf{Config} 
  & \multicolumn{2}{c}{\textbf{RLTD}} 
  & \multicolumn{2}{c}{\textbf{MU3D}} \\
\cmidrule(lr){4-5}\cmidrule(lr){6-7}
 & & & \textbf{Acc} & \textbf{F1} & \textbf{Acc} & \textbf{F1} \\
\midrule

% -----------------------
% Llama-3.1-8B
% -----------------------
\multirow{6}{*}{LLaMA 3.1} 
 & \multirow{2}{*}{\textit{nv}}
   & \textit{zs}
     & 51.24 & 35.34 
     & - & -  \\
 & 
   & \textit{\textit{fs}}
     & 63.02 & 62.79
     & - & -  \\
\cmidrule(lr){2-7}
 & \multirow{2}{*}{\textit{vs}}
   & \textit{zs}
     & 52.07 & 38.40
     & 50.63 & 49.67 \\
 &
   & \textit{\textit{fs}}
     & 62.60 & 61.72 
     & 51.88 & 49.44 \\
\cmidrule(lr){2-7}
 & \multirow{2}{*}{\textit{as}}
   & \textit{zs}
     & 57.85 & 57.56
     & 49.06 & 49.00 \\
 &
   & \textit{\textit{fs}}
     & 64.74 & 63.12
     & 51.57 & 51.25 \\
\midrule

% -----------------------
% Gemma2-9B
% -----------------------
\multirow{6}{*}{Gemma 2} 
 & \multirow{2}{*}{\textit{nv}}
   & \textit{zs}
     & 52.07 & 38.40 
     & - & -  \\
 & 
   & \textit{\textit{fs}}
     & 66.67 & 64.85
     & - & -  \\
\cmidrule(lr){2-7}
 & \multirow{2}{*}{\textit{vs}}
   & \textit{zs}
     & 52.07 & 38.40 
     & 51.25 & 44.16 \\
 &
   & \textit{\textit{fs}}
     & 66.94 & 66.92 
     & \textbf{53.44} & \textbf{51.56} \\
\cmidrule(lr){2-7}
 & \multirow{2}{*}{\textit{as}}
   & \textit{zs}
     & 66.94 & 66.28
     & 49.38 & 45.82 \\
 &
   & \textit{\textit{fs}}
     & 68.60 & 68.59
     & 51.56 & 50.98 \\
\midrule

% -----------------------
% GPT-4o
% -----------------------
\multirow{6}{*}{GPT-4o} 
 & \multirow{2}{*}{\textit{nv}}
   & \textit{zs}
     & 65.29 & 61.71
     & - & -  \\
 & 
   & \textit{\textit{fs}}
     & \textbf{72.93} & \textbf{72.84} 
     & - & -  \\
\cmidrule(lr){2-7}
 & \multirow{2}{*}{\textit{vs}}
   & \textit{zs}
     & 65.29 & 65.00
     & 52.19 & 47.98 \\
 &
   & \textit{\textit{fs}}
     & 69.42 & 69.00 
     & 54.69 & 52.48 \\
\cmidrule(lr){2-7}
 & \multirow{2}{*}{\textit{as}}
   & \textit{zs}
     & 66.12 & 64.73
     & 49.38 & 42.31 \\
 &
   & \textit{\textit{fs}}
     & 67.69 & 66.85
     & 51.63 & 45.21 \\
\bottomrule
\end{tabular}
}
\captionsetup{font=scriptsize}
\caption{Comparison of auxiliary features for the multimodal datasets. \textit{zs}: zero shot; \textit{\textit{fs}}: few shot; \textit{nv}: non-verbal; \textit{vs}: video summaries, \textit{as}: audio summaries.}
\label{tab:aux_features_multimodal}
\end{table}

In Table~\ref{tab:aux_features_multimodal}, we compare three types of auxiliary features: non-verbal gestures, LLM-generated audio and video summaries under both zero-shot and few-shot settings. Each model uses randomly selected in-context examples when operating in the few-shot configuration. From these results, we observe that including non-verbal gestures alongside the transcript yields a modest improvement for GPT-4o ($\sim1.4$ points in F1 score on the RLTD dataset). This gain is consistent with GPT-4o’s demonstrated strengths in in-context learning. In contrast, the inclusion of non-verbal features negatively impacts LLaMA 3.1 and Gemma 2: their tendency to overpredict the \texttt{Deceptive} label suggests that limited in-context examples are insufficient for these models to learn patterns of non-verbal gestures across truthful and deceptive scenarios. Turning to video summaries, GPT-4o again exhibits relative gains on MU3D, although the improvements for other models and datasets remain negligible or even degrade performance. A similar pattern holds for audio summaries: while certain configurations see a slight boost, many are on par with or slightly below the corresponding transcript-only results. 
Overall, additional features do not universally enhance predictive accuracy without fine-tuning.
\section{Conclusion}
Our comprehensive evaluation reveals that LLMs and LMMs exhibit promising capabilities for deception detection across diverse contexts. While fine-tuning significantly enhanced performance, improvements on multimodal datasets are still lagging, highlighting persistent challenges in capturing nuanced cross-modal deception cues in LMMs. Moreover, incorporating reasoning generation to explain predictions did not consistently improve overall accuracy over straightforward label prediction, emphasizing that the inherently ambiguous nature of deception cues makes it harder for the models to reason successfully. These findings underscore the importance of careful prompt design and in-context example selection while pointing to the need for further methodological refinements in practical deception detection applications.

\section*{Limitations}
While our study shows promising results, it has several limitations that pave the way for future research. First, our experiments are limited to English-language datasets, which may not fully capture the linguistic diversity or cultural nuances necessary for broader applicability. Second, we focus exclusively on human deception, leaving the detection of AI-generated deceptive behaviors as an area for further exploration. Third, the reliance on a limited range of publicly available datasets and controlled scenarios may affect the generalizability of our findings to more varied, real-world contexts. Additionally, the deployment of deception detection systems involves ethical, privacy, and interpretability challenges that must be carefully managed, especially in legal or interpersonal settings. Finally, the computational cost—approximately $300$ USD for experiments with GPT-based models—and the significant GPU resources required for open-source models highlight practical considerations for real-world implementation.

\makeatletter
\ifacl@finalcopy
\@ifpackagewith{acl}{review}{}{
\section*{Acknowledgements}
We would like to thank the anonymous reviewers for their valuable feedback and input. We gratefully acknowledge support from National Science Foundation via the award IIS-1942918 as well as support from the Texas A\&M Institute of Data Science via an internal grant. Portions of this research were conducted with the advanced computing resources provided by Texas A\&M High-Performance Research Computing.
}
\else
\fi
\makeatother

\appendix

\section{Models}

All the models except GPT-4o are hosted on huggingface. GPT-4o model is used via OpenAI API.

\begin{table*}[ht]
\centering
\small
\begin{tabular}{|l|l|l|}
\hline
\textbf{Model Name} & \textbf{Model ID} & \textbf{License}\\ \hline
LLaMA 3.1 & \texttt{meta-llama/Llama-3.1-8B-Instruct} & llama 3.1             \\ \hline
Gemma 2 & \texttt{google/gemma-2-9b-it}  & gemma            \\ \hline
GPT-4o & \texttt{gpt-4o-2024-08-06} & proprietary \\ \hline
LLaVA-NEXT-Video & \texttt{llava-hf/LLaVA-NeXT-Video-7B-hf} & llama2             \\ \hline
Qwen2VL & \texttt{Qwen/Qwen2-VL-7B-Instruct} & apache-2.0 \\ \hline
MERaLiON-AudioLLM & \texttt{MERaLiON/MERaLiON-AudioLLM-Whisper-SEA-LION} & meralion-public-license \\ \hline
Qwen2-Audio & \texttt{Qwen/Qwen2-Audio-7B-Instruct} & apache-2.0 \\ \hline
\end{tabular}
\caption{Model Information}
\label{tab:model-info}
\end{table*}

\section{List of Non-verbal Features}
\label{non-verbal}
RLTD dataset comes with a set of 40 manually annotated non-verbal features. These features are broadly categorized into \textit{facial displays} and \textit{hand movements}. The original annotation provides a binary value for each of these features with respect to whether these attributes were demonstrated by the primary speaker in the video. We filter the most relevant 16 features and use the feature names directly for generating LLM predictions. A list of these features is - \texttt{Both Hands Movement, Complex Hands Trajactory, Downwards Lip Movement, Eyes Closing Repeatedly, Frown, Gaze Down, Gaze Side, Gaze at Interlocutor, Head Down, Mouth Closed, Mouth Opened, Raise Eyebrows, Repeated Nods, Scowl, Single Hand Movement, Upwards Lip Movement}.

\section{Prompts}
\subsection{Prompt for Direct Label Prediction}

\texttt{You are an expert judge, specializing in deception detection. Your task is to evaluate transcripts from courtroom trials or interviews with exceptional accuracy. Based on the provided transcript, assess whether the speaker's statements are 'Truthful' or 'Deceptive'. Utilize linguistic cues, contextual clues, and logical consistency in the transcript to make your judgment. Respond with only one word: either 'Truthful' or 'Deceptive'. Do not provide additional explanation, reasoning, or commentary in your response.} 

\texttt{Below are some reference examples: \textbf{<few shot examples>}}

\texttt{Carefully analyze the transcript below, paying close attention to the content and context, and provide your judgment.}

\subsection{Prompt for Reasoning Generation}
\texttt{You are an expert judge, specializing in deception detection. Your task is to evaluate transcripts from courtroom trials or interviews with exceptional accuracy. Based on the provided transcript, assess whether the speaker's statements are 'Truthful' or 'Deceptive'. Utilize linguistic cues, contextual clues, and logical consistency in the transcript to make your judgment. Respond with both your judgment and the reasoning behind your judgment. Provide your response in the format- Judgment: <your judgment in one word 'Truthful' or 'Deceptive'>, Reasoning: <your reasoning for the judgment based on the transcript and your analysis>.}

\texttt{Below are some reference examples: \textbf{<few shot examples>}}

\texttt{Carefully analyze the transcript below, paying close attention to the content and context, and provide your answer in the desired format.}

\subsection{Prompt for Video Summary Generation}
\label{subsec:vid-sum-prompt}
\texttt{Analyze this video and provide a comprehensive and detailed summary based on:
- Speaker details (appearance, positioning); if there are more than one person, focus only on the person being interviewed or who is answering questions.
- Facial expressions (micro-expressions, eye movements, eye-brow movements, lip and mouth movements etc.)
- Body language (hand gestures, head movements, posture changes etc.)
- Physical indicators (visible stress signs, fidgeting patterns etc.)
- Key scene descriptions (describing the most crucial moments from the video)
Describe any notable behavioral patterns or changes over time. Focus on any observable visual cues. The final summary should be a paragraph containing all the important information extracted from the input video according to the instructions provided. 
}

\subsection{Prompt for Audio Summary Generation}
\texttt{Analyze the input audio and provide a summary of the pitch and tone of the speaker in the audio recording. Describe any notable acoustic patterns briefly.
}

\section{Response Generation Strategies}
\label{reasoning gen}
\begin{table*}[ht]
\centering
\small
% \resizebox{\columnwidth}{!}{
\begin{tabular}{l c c cc cc cc}
\toprule
\textbf{LLM} & \textbf{Config} & \textbf{Response}
  & \multicolumn{2}{c}{\textbf{RLTD}}
  & \multicolumn{2}{c}{\textbf{MU3D}}
  & \multicolumn{2}{c}{\textbf{OpSpam}} \\
\cmidrule(lr){4-5}\cmidrule(lr){6-7}\cmidrule(lr){8-9}
 & & \textbf{Generation} & \textbf{Acc} & \textbf{F1}
     & \textbf{Acc} & \textbf{F1}
     & \textbf{Acc} & \textbf{F1} \\
\midrule

% -----------------------
% Llama-3.1-8B
% -----------------------
\multirow{4}{*}{LLaMA 3.1} 
 & \multirow{2}{*}{zero shot} 
   & label 
     &  54.67 & 51.00
     &  49.61 & 48.51
     &  52.35 & 52.33 \\
 & 
   & label + reasoning 
     & 52.07 & 50.27 
     & 48.44 & 47.95 
     & 51.18 & 51.17 \\
\cmidrule(lr){2-9}
 & \multirow{2}{*}{few shot}
   & label
     & 68.87 & 68.14 
     & 51.72 & 51.18 
     & 59.62 & 59.19 \\
 &
   & label + reasoning
     & 65.71 & 65.61
     & \textbf{56.49} & \textbf{56.15} 
     & 61.43 & 60.83 \\
\midrule

% -----------------------
% Gemma2-9B
% -----------------------
\multirow{4}{*}{Gemma 2} 
 & \multirow{2}{*}{zero shot} 
   & label 
     & 67.77 & 66.67
     & 52.35 & 48.20 
     & 49.28 & 48.45 \\
 & 
   & label + reasoning
     & 66.12 & 64.20
     & \underline{55.63} & \underline{54.42}
     & 50.28 & 47.00 \\
\cmidrule(lr){2-9}
 & \multirow{2}{*}{few shot}
   & label
     & \underline{69.63} & \underline{69.52} 
     & 54.22 & 52.34 
     & 57.70 & 57.59 \\
 &
   & label + reasoning
     & 68.18 & 68.00 
     & 53.91 & 50.73
     & 59.68 & 57.75 \\
\midrule

% -----------------------
% GPT-4o
% -----------------------
\multirow{4}{*}{GPT-4o} 
 & \multirow{2}{*}{zero shot}
   & label 
     & 67.63 & 67.62 
     & 52.42 & 43.98
     & 58.53 & 53.72 \\
 & 
   & label + reasoning
     & 64.46 & 64.31
     & 51.41 & 41.89 
     & 59.04 & 53.63\\
\cmidrule(lr){2-9}
 & \multirow{2}{*}{few shot}
   & label
     & \textbf{71.69} & \textbf{71.39} 
     & 53.20 & 46.86 
     & \textbf{68.40} & \textbf{67.58} \\
 &
   & label + reasoning
     & \underline{69.63} & 69.07 
     & 52.17 & 42.67
     & \underline{64.20} &  \underline{61.04}\\
\bottomrule
\end{tabular}
% }
\captionsetup{font=footnotesize}
\caption{Comparison of different response generation strategies (direct label prediction vs.\ post-hoc reasoning generation) under zero-shot and few-shot settings. The few-shot examples are randomly selected. The best and second best results are indicated by \textbf{bold} and \underline{underline} respectively.}
\label{tab:response_generation_shots}
\end{table*}

The results in Table~\ref{tab:response_generation_shots} offer a comparative analysis between direct label prediction and post-hoc reasoning generation across the three datasets. We systematically evaluate whether generating reasoning after the label contributes positively to the model’s predictive performance, under both zero-shot and few-shot prompting settings. Across most of the settings, particularly on RLTD dataset, direct label prediction tends to yield higher accuracy and F1 scores. For example, GPT-4o achieves an F1 score of 71.39 on RLTD with few-shot direct label prediction, outperforming its label+reasoning counterpart (69.63 F1 score). However, this trend does not hold universally. In the MU3D dataset, which involves scripted deception, post-hoc reasoning occasionally matches or slightly improves performance. LLaMA 3.1, for instance, reaches its best F1 score of 56.15 on MU3D using few-shot post-hoc reasoning. In the OpSpam dataset, dominated by textual content, the advantage again leans toward direct label prediction. GPT-4o in particular shows a noticeable drop in F1 score from 67.58 (few-shot label) to 61.04 (few-shot label + reasoning), suggesting that the inclusion of generated explanations may introduce noise or ambiguity, especially when no visual or behavioral cues are available to ground the reasoning. While post-hoc reasoning generation provides interpretability of model predictions, it does not consistently improve classification performance, and in many cases, leads to modest degradation. 

\section{Performance Trends in Few-Shot Learning}
\label{sec:llm_nshots}
\begin{figure*}
    \centering
    \includegraphics[width=0.90\textwidth]{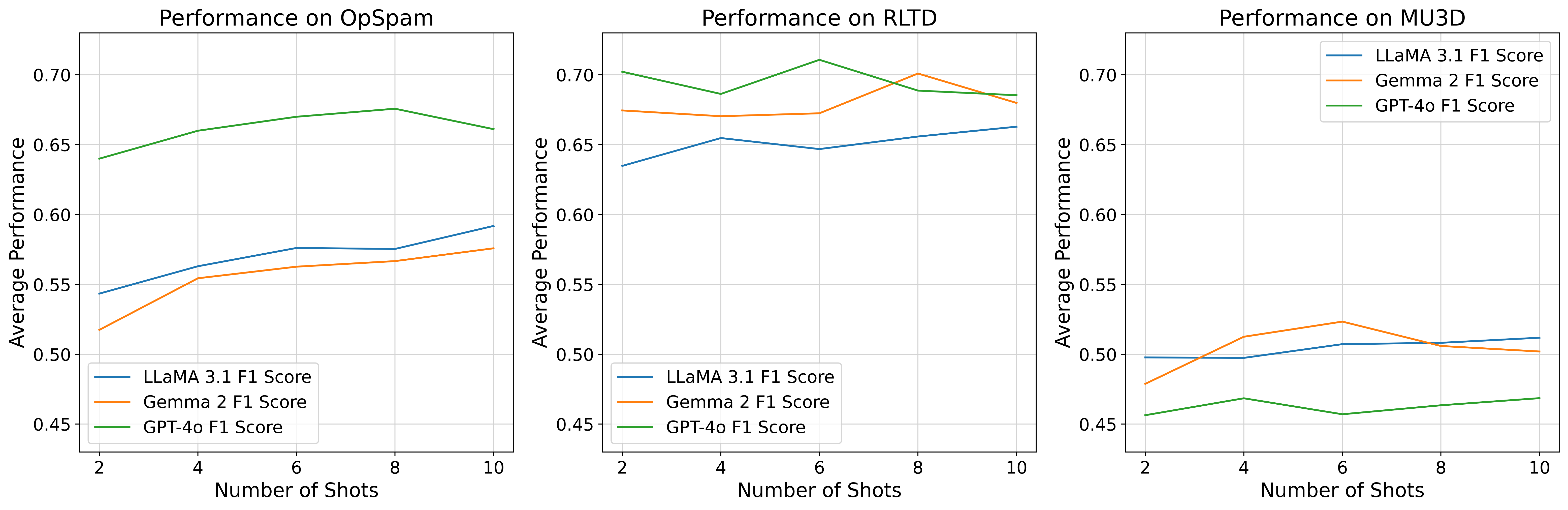}
    \captionsetup{font=scriptsize}
    \caption{F1 score across n-shots in few shot learning}
    \label{fig:llm_shots}
\end{figure*}

In our study, we examined how large language models (LLMs) perform across various datasets using few-shot prompting. We calculated the F1 score by averaging results across all seeds to ensure consistent measurement as illustrated in Figure ~\ref{fig:llm_shots}. Our findings reveal that models like GPT-4o initially improve with more few-shot examples, demonstrating their ability to use additional data effectively. However, this improvement subsequently declined when too many examples were provided, likely due to the increased complexity of the prompts complicating the model's reasoning ability. LLaMA 3.1 consistently showed significant gains with an increased number of examples for OpSpam and MU3D, indicating strong adaptability to more extensive data inputs. Gemma 2's performance improved on the OpSpam dataset with more examples but declined on the MU3D and RLTD datasets after a certain point. This pattern suggests a possible optimization ceiling, where additional examples no longer contribute to performance enhancements and may instead hinder the model's effectiveness due to prompt saturation.

\begin{figure}
    \centering
    \includegraphics[width=0.5\textwidth]{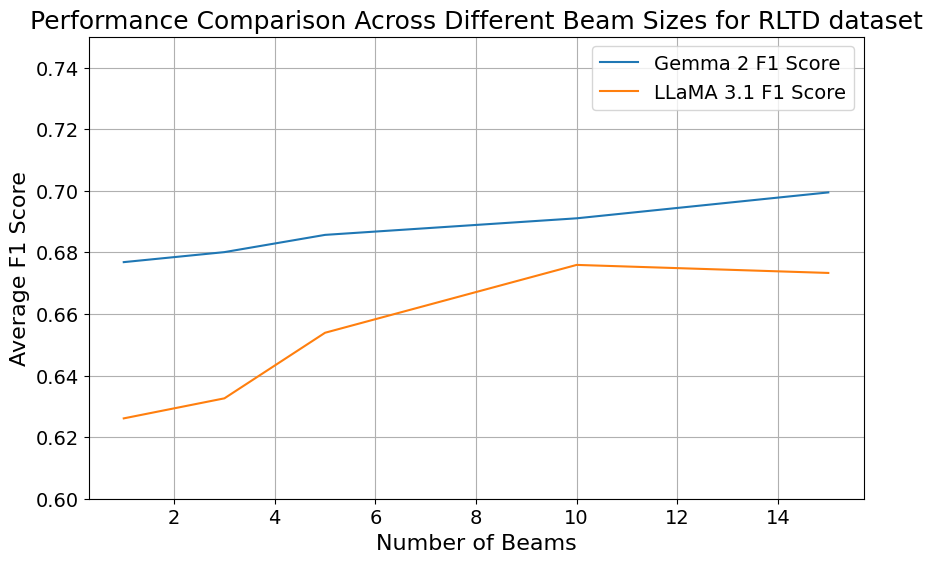}
    \captionsetup{font=scriptsize}
    \caption{F1 score across different beam sizes on RLTD dataset}
    \label{fig:llm_beams}
\end{figure}

\section{Evaluating the Efficacy of Beam Search in Reasoning}
\label{sec:llm_beams}

We performed beam search to evaluate its potential to enhance performance in label generation with reasoning for the LLaMA and Gemma models on the RLTD dataset. We calculated the average F1 scores for 10-shot experiment across all seeds for varying beam sizes, specifically [1, 3, 5, 10, 15], with the results detailed in Figure ~\ref{fig:llm_beams}. Our analysis reveals that as the number of beams increases, the performance generally increases. However, even the best outcomes obtained through the beam search failed to surpass the performance levels achieved via direct label generation for the RLTD dataset.

\section{LLM Reasoning Analysis Examples}
\begin{figure*}
    \centering
    \includegraphics[width=1\textwidth]{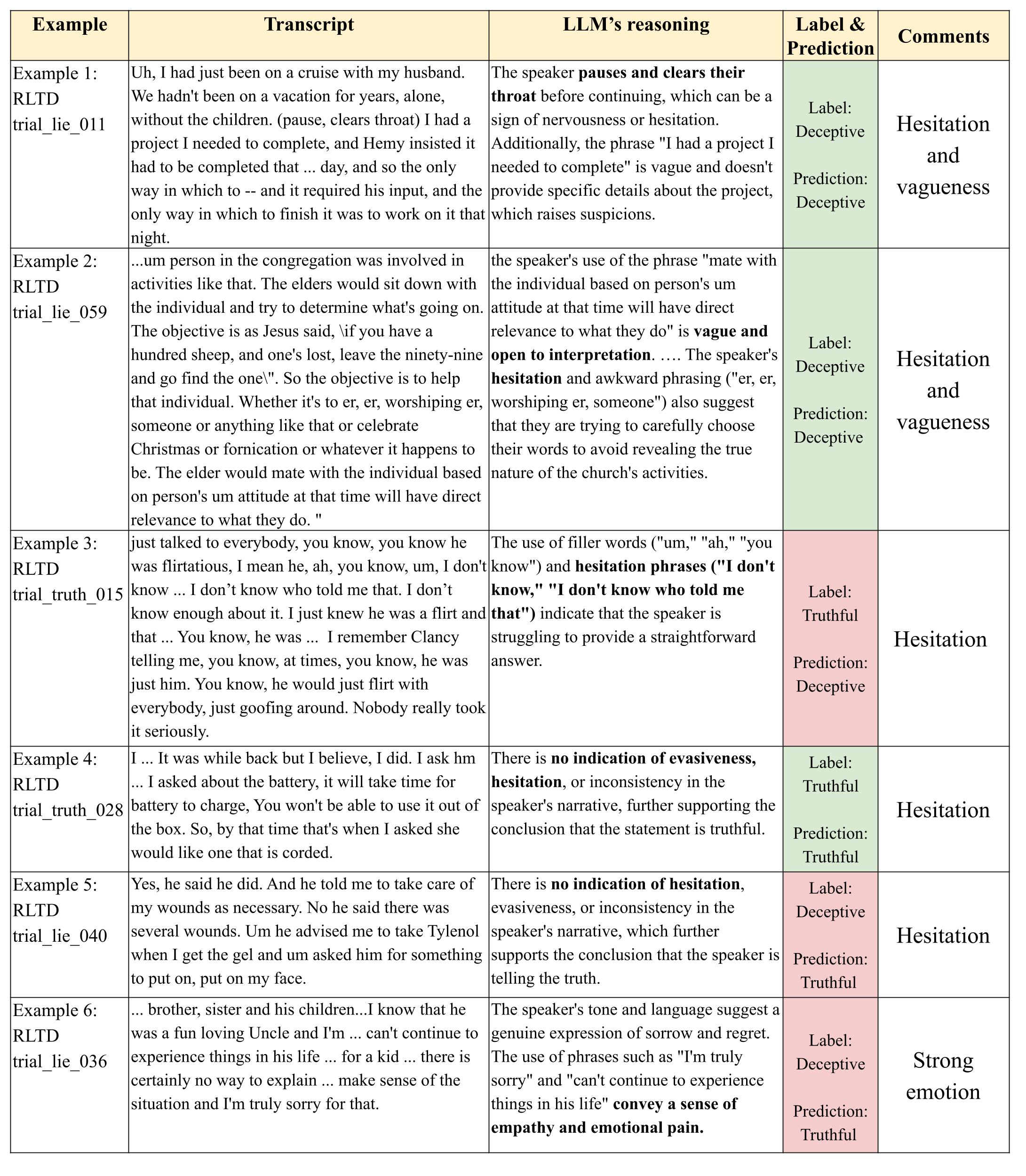}
    \captionsetup{font=scriptsize}
    \caption{Examples of LLM Reasoning on RLTD Dataset}
    \label{fig:llm_reason1}
\end{figure*}

\begin{figure*}
    \centering
    \includegraphics[width=1\textwidth]{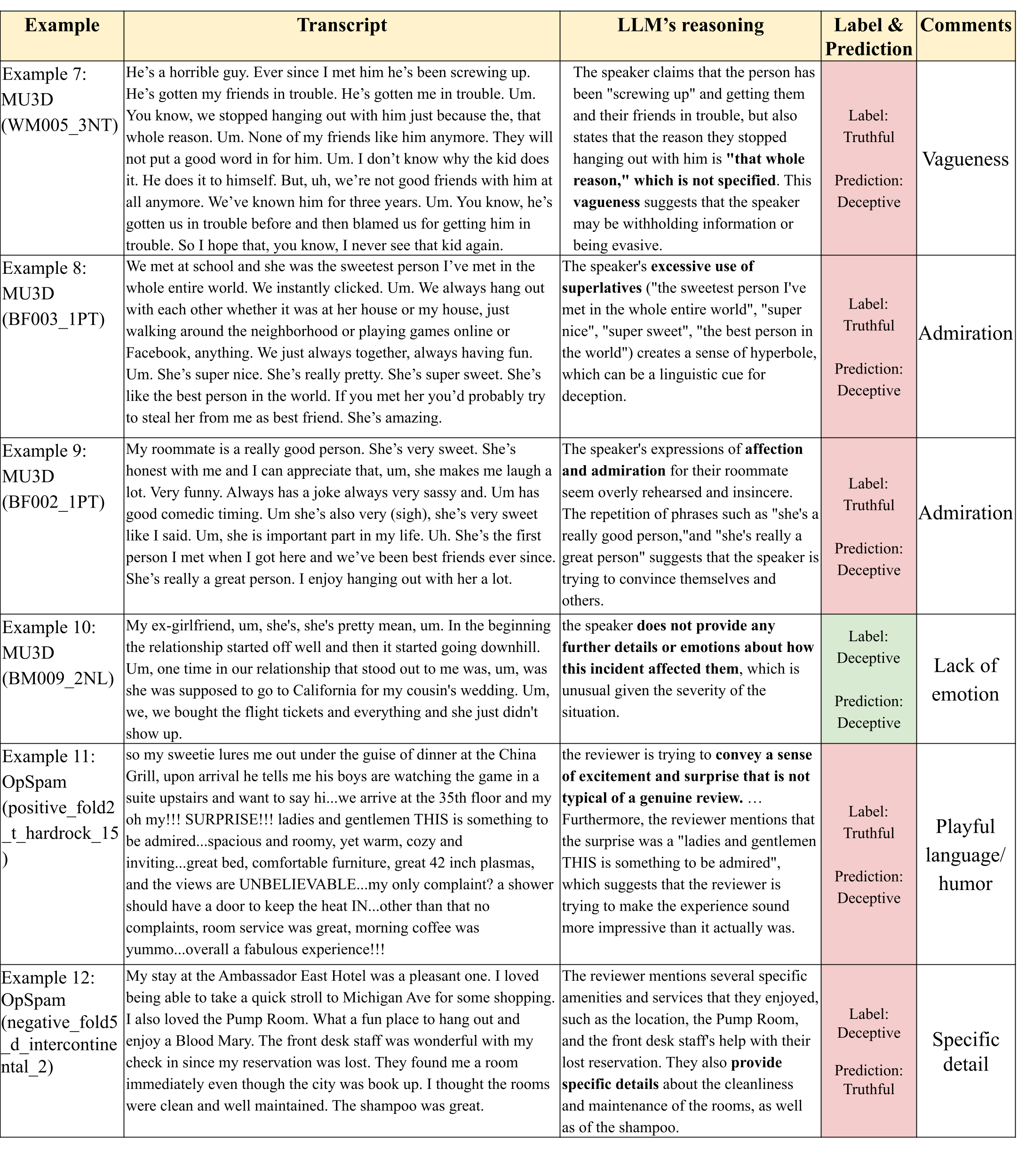}
    \captionsetup{font=scriptsize}
    \caption{Examples of LLM Reasoning on MU3D and OpSpam Dataset}
    \label{fig:llm_reason2}
\end{figure*}

We have illustrated several examples on the basis of deception cues across three datasets in  Figure \ref{fig:llm_reason1}, Figure \ref{fig:llm_reason2}.
\label{sec:llm_ra}

\section{Post-hoc Reasoning Generation vs Chain-of-Thought Reasoning}
\label{cot}
\begin{table}[ht]
\centering
\small
\resizebox{\columnwidth}{!}{
\begin{tabular}{l c cc cc cc}
\toprule
% -- First header row --
\textbf{LLM} 
  & \textbf{Order}
  & \multicolumn{2}{c}{\textbf{RLTD}}
  & \multicolumn{2}{c}{\textbf{MU3D}}
  & \multicolumn{2}{c}{\textbf{OpSpam}} \\
\cmidrule(lr){3-4}\cmidrule(lr){5-6}\cmidrule(lr){7-8}
% -- Second header row --
 & 
 & \textbf{Acc} & \textbf{F1}
 & \textbf{Acc} & \textbf{F1}
 & \textbf{Acc} & \textbf{F1} \\
\midrule

% ---- LLaMA 3.1 ----
\multirow{2}{*}{LLaMA 3.1} 
  & \textit{l $\rightarrow$ r}    & 65.71 & 65.61 & 56.49 & 56.15 & 61.43 & 60.83 \\
  &\textit{r $\rightarrow$ l}    & 58.13 & 55.31 & 52.08 & 49.50 & 58.60 & 57.62 \\
\midrule

% ---- Gemma 2 ----
\multirow{2}{*}{Gemma 2} 
  & \textit{l $\rightarrow$ r}    & 68.18 & 68.00 & 53.91 & 50.73 & 59.68 & 57.75 \\
  & \textit{r $\rightarrow$ l}    & 58.68 & 54.39 & 50.63 & 50.39 & 61.62 & 59.36 \\
\bottomrule
\end{tabular}
}
\captionsetup{font=scriptsize}
\caption{Comparison of LLM performances under different reasoning generation orderings across datasets. \textit{l $\rightarrow$ r}: label $\rightarrow$ reasoning; \textit{r $\rightarrow$ l}: reasoning $\rightarrow$ label.}
\label{tab:cot}
\end{table}

While generating additional reasoning for predicted labels, we adopt a \textbf{post-hoc reasoning generation} strategy, where the model first outputs the classification label, followed by a justification. An alternative is \textbf{chain-of-thought reasoning}, where the model first reasons over the input before predicting the final label. We chose the post-hoc approach based on empirical evidence. Specifically, we conducted a controlled comparison of the two prompting strategies — (a) label $\rightarrow$ reasoning (post-hoc) and (b) reasoning $\rightarrow$ label (chain-of-thought)—under a few-shot setting across three datasets. As shown in Table~\ref{tab:cot}, the post-hoc reasoning generation strategy generally outperforms chain-of-thought.

\begin{table*}[ht]
\centering
\small  % or \footnotesize
% \resizebox{\columnwidth}{!}{
\begin{tabular}{l|l|c|c|c}
\toprule
\textbf{Model Type} & \textbf{Model} & \textbf{Config} & \textbf{Accuracy} & \textbf{F1} \\
\midrule
\multirow{3}{*}{Baselines} & RoBERTa-ft & - & 73.16 & 72.88 \\
& BiLSTM+Attention & - & 71.44 & 71.19 \\
& CNN & - & 69.92 & 69.70 \\
\hline
\multirow{6}{*}{LLM Inference} & LLaMA 3.1 & zero shot & 59.76 & 59.72 \\
& LLaMA 3.1 & few shot & 64.32 & 64.16 \\
& Gemma 2 & zero shot & 57.31 & 55.95 \\
& Gemma 2 & few shot & 62.36 & 61.95 \\
& Gpt-4o & zero shot & 64.19 & 64.10 \\
& Gpt-4o & few shot & 64.41 & 64.17 \\
\hline
\multirow{2}{*}{LLM Finetuning} & LLaMA 3.1 & - & \underline{73.40} & \underline{73.31} \\
& Gemma 2 & - & \textbf{73.55} & \textbf{73.54} \\
\bottomrule
\end{tabular}
% }
\captionsetup{font=footnotesize}
\caption{LLM and baseline model performances Comparison on Prolific dataset.}
\label{tab:prolific}
\end{table*}

In addition to the performance improvements, the post-hoc reasoning reasoning strategy offers better interpretability for analyzing LLM's decision making process. When reasoning is generated prior to the label, LLMs often produce explanations that weigh cues supporting both possible classes. This makes it difficult to use the generated reasoning to interpret the model’s decision, as it lacks a clear alignment with the predicted label. An example of LLM prediction through chain-of-thought -
\\
\texttt{Reasoning: 1. The speaker starts with "Ah, I don't remember," which is a common phrase used when someone is unsure or trying to downplay their memory of an event. This could be a sign of deception, but it's not conclusive on its own.
2. The speaker then says "evidently I, I -- I tried to call Hemy," which is a vague statement. The use of "evidently" suggests that the speaker is not entirely sure about their actions, and the repetition of "I" creates a sense of hesitation.
[...] 6. Overall, the speaker's statements seem inconsistent and vague, which could be a sign of deception. However, the timeline could suggest truthfulness.
\\
Judgment: Deceptive}

In this case, the reasoning incorporates both pro-truthful and pro-deceptive cues before arriving at a final judgment. Hence, it becomes challenging to systematically identify which cues LLMs rely on most frequently for decision-making at scale. In contrast, the post-hoc reasoning generation strategy leads to more focused, label-aligned justifications that are easier to interpret and trace back to specific cues - 
\\
\texttt{Judgment: Deceptive
\\
Reasoning: The speaker claims they tried to call Hemy (presumably their boss) to inform them that they had left the office and something had happened to their husband. However, the speaker immediately follows this statement with "Which seemed -- was a very normal thing for me to have done." This phrase suggests an attempt to justify or rationalize their actions, which is a common linguistic cue for deception. Additionally, the speaker's initial claim "I don't remember" followed by a detailed account raises a red flag, indicating they may be filling in memory gaps with plausible details.
}

Based on both empirical evidence and interpretability advantages, we adopt the post-hoc reasoning generation approach over chain-of-thought prompting.

\section{Data-Origin Confound in OpSpam Dataset}
\label{sec:opspam}
OpSpam is one of the most widely used benchmarks in textual deception detection and thus provides a relevant testbed within the scope of our study. However, contrasting Mechanical Turk-generated deceptive reviews with scraped genuine ones introduces a data-origin confound, potentially inflating model performance by encouraging reliance on stylistic artifacts rather than true deception cues, as discussed in ~\citet{crowdsourcedvsreal, Soldner_2022}. Our primary goal is to evaluate the behavior of LLMs and LMMs under zero and few-shot settings with limited in-domain supervision. As shown in Table \ref{tab:response_generation_shots}, LLMs don't appear to exploit OpSpam dataset bias in the zero-shot setup because they cannot infer such confounding factors from single input data points. In the case of few-shot results with randomly selected examples, we do observe a performance improvement from zero-shot to few-shot for OpSpam but that is consistent with other datasets. For instance, LLaMA 3.1 on RLTD experiences a 17.14\% improvement, whereas the imrpovement on the OpSpam dataset is 6.86\%. However, Table \ref{tab:example_selection} reveals that when using semantically similar (sim-top) few-shot examples, the average performance improvement across 3 models is 5.54\% over random example selection in the OpSpam dataset. This gain is higher than that of RLTD (3.19\%) and MU3D (4.69\%), which suggests that with as few as 10 carefully curated in-context examples, LLMs may begin to pick up on underlying dataset-specific patterns, including potential biases. Understanding how LLMs leverage these biases offers valuable insights for designing more robust deception detection systems. To expand our evaluation beyond crowdsourced reviews, we have conducted additional experiments with the dataset from Confounds and Overestimations in Fake Review Detection~\citep{Soldner_2022}, specifically under its “Pure Veracity” setting. this setting is particularly challenging since both the truthful and deceptive reviews are coming from real-world owners of the smartphones gathered via the Prolific platform. This setting is particularly challenging since both the truthful and deceptive reviews are coming from real-world owners of the smartphones gathered via the Prolific platform. The results on Prolific dataset using our text-based baselines and the same LLM approaches as discussed in the paper, are presented in Table~\ref{tab:prolific}. Experimental results indicate that fine-tuned model performance drops notably compared to OpSpam, but the overall comparative trend remains similar, with fine-tuned LLaMA 3.1 and Gemma 2 models outperforming the baselines. In zero and few-shot setups, we observe similar performance on both OpSpam and the Prolific dataset, further confirming, LLMs cannot pick up the nuanced platform-specific biases very well with limited in-domain examples.

\section{Hyper-parameters and Budgeting}
\subsection{Baselines}
We use a learning rate of 4e-5 for training the baselines and the models are trained for 20 epochs. The models are trained on 1 A6000 GPU.

\subsection{LLMs}
For few-shot examples, we explore 2, 4, 6, 8, 10 examples and report the best results for few shot performance. All the results reported are an average of 3 seeds.

\section{AI Assistance}
\label{sec:aia}
We have used ChatGPT for writing assistance in the paper writing.
\end{document}